%% file: tcyb2017_exposureFusion.tex
\documentclass[journal]{IEEEtran}
\input{def}
\usepackage{booktabs}
\usepackage{amsmath} 
\usepackage{amssymb}
\usepackage{multirow} 

\usepackage{marvosym}
\usepackage[ruled]{algorithm2e} 

\usepackage{url}
\usepackage{mathtools}
\ifCLASSINFOpdf
\else
\fi
 \usepackage[caption=false,font=footnotesize]{subfig} 
\begin{document}
%
\title{
A Bio-Inspired Multi-Exposure Fusion Framework for Low-light Image Enhancement 
}
%
%
%

\author{Zhenqiang~Ying,~\IEEEmembership{Student Member,~IEEE,}
        Ge~Li,~\IEEEmembership{Member,~IEEE,}
        and~Wen~Gao,~\IEEEmembership{Fellow,~IEEE}
\thanks{
This work was supported by the grant of National Science Foundation of China (No.U1611461), Shenzhen Peacock Plan (20130408-183003656), and Science and Technology Planning Project of Guangdong Province, China (No. 2014B090910001). 
This paper was recommended by Associate Editor X. XX.
}
\thanks{Z. Ying, G. Li, and W. Gao are with the School of 
Electronic and Computer Engineering, Shenzhen Graduate School, Peking University, 
518055 Shenzhen, China (e-mail:
zqying@pku.edu.cn; geli@ece.pku.edu.cn; wgao@pku.edu.cn).
}
\thanks{Color versions of one or more of the figures in this paper are available
online at http://ieeexplore.ieee.org.
}
\thanks{Digital Object Identifier XX.XXXX/TCYB.20XX.XXXXXXX
}
\thanks{Manuscript received XXX XX, 20XX; revised XXX XX, 20XX.}
}

%
%

\markboth{Journal of \LaTeX\ Class Files,~Vol.~14, No.~8, August~2015}%
{Shell \MakeLowercase{\textit{et al.}}: Bare Demo of IEEEtran.cls for IEEE Journals}
%



\maketitle

\begin{abstract}
Low-light images are not conducive to human observation and computer vision algorithms due to their low visibility. 
Although many image enhancement techniques have been proposed to solve this problem, 
existing methods inevitably introduce contrast under- and over-enhancement.
Inspired by human visual system, we design a multi-exposure fusion framework for low-light image enhancement.
Based on the framework, we propose a dual-exposure fusion algorithm to provide an accurate contrast and lightness enhancement.
Specifically, we first design the weight matrix for image fusion using illumination estimation techniques. 
Then we introduce our camera response model to synthesize multi-exposure images.
Next, we find the best exposure ratio so that the synthetic image
is well-exposed in the regions where the original image is under-exposed.
Finally, the enhanced result is obtained by fusing the input image and the synthetic image according to the weight matrix.
Experiments show that our method can obtain results with less contrast and lightness distortion compared to that of several state-of-the-art methods.
\end{abstract}


%
\IEEEpeerreviewmaketitle

\section{Introduction}



\IEEEPARstart{W}{ITH}
the development of photographic techniques, the image quality is 
greatly improved in both resolution and bit-depth.
However, images captured by standard imaging devices often suffer from low visibility in non-uniform illuminated environments such as back lighting, nighttime and low-light indoor scene.
Those images may lose information in under-exposed regions, making the image content invisible to human eyes.
Since the camera dynamic range is limited, 
if we increase camera exposure to reveal the information of under-exposed regions, the well-exposed regions will be over-exposed or even saturated.
To address the problem,
many image enhancement techniques have been proposed
including histogram-based methods \cite{xu2014generalized,celik2014spatial,lee2012ldr,arya2013histogram,lee2012power,celik2011cvc,arici2009wahe,coltuc2006exact,chen2003rmshe,chen2003mmbebhe,pizer1987adaptive,pisano1998contrast,kim1997contrast}
, Retinex-based methods~\cite{guo2017lime,wang2014variational,wang2013naturalness,kimmel2003variational,jobson1997properties,ng2011total,jobson1997multiscale},
Logarithmic Image Processing methods~\cite{panetta2011parameterized,panetta2008human},
 and filtering-based methods~\cite{wang2013automatic,yuan2012automatic,deng2011generalized,shan2010globally,hong2009novel}.
Although some methods can obtain results with good subjective quality, those results may not accurately reflect the true lightness and contrast of the scene.
So,
accurate light and contrast enhancement based on a single image is still a challenging problem.

\begin{figure}[t]
\centering
\subfloat[]{\includegraphics[width=0.8in]{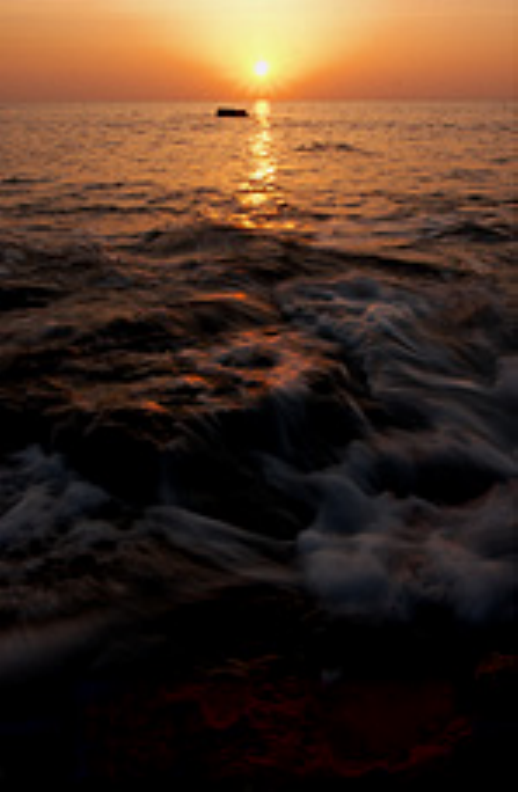} }\hfil
\subfloat[]{\includegraphics[width=0.8in]{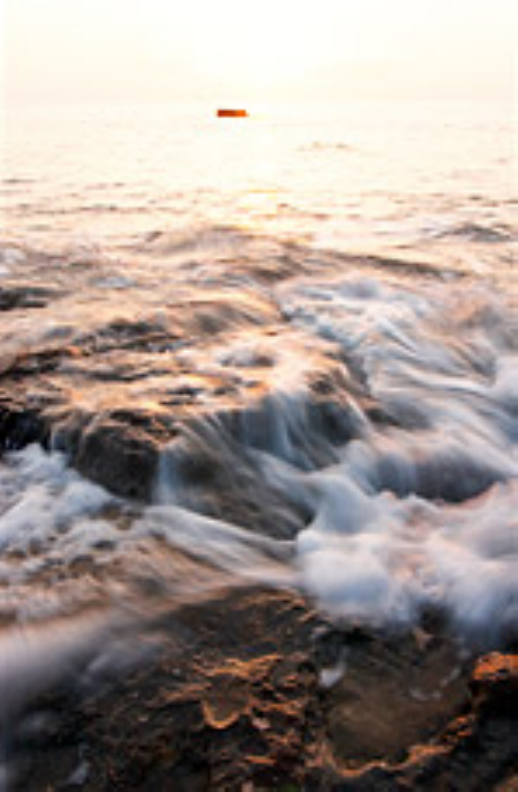} }\hfil
\subfloat[]{\includegraphics[width=0.8in]{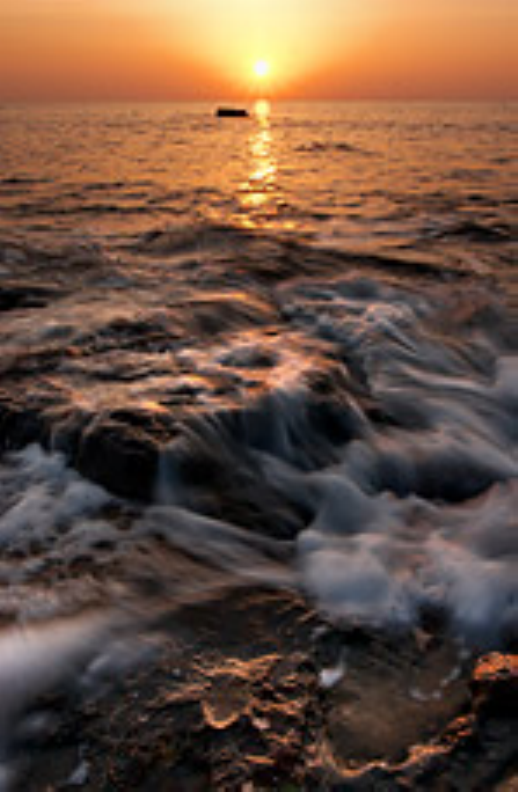}}
\caption{
Our mental image is a fused version of different eye exposure. (dynamically adjusting).
(a) Eye focuses on background.
(b) Eye focuses on foreground.
(c) Our mental image.
\cite{bookUnderstandingPhotography}}
\label{fig:bio}
\end{figure}

With a set of different exposure images in the same scene, High Dynamic Range (HDR) techniques can synthesize images that are close to the perceived scene. 
We know that the camera and the human eye have a lot of similarities, 
then, why we can perceive an image that is well-exposed everywhere while the camera cannot? 
%
The reason lies in the post-processing in our brains that have an image fusion mechanism similar to HDR technique \cite{bookUnderstandingPhotography}.
Human eye exposure changes with its focus point, resulting in 
multi-exposure image set which is then sent to the brain.
Although each one of those images suffers from under-exposed or over-exposed problem in some regions,
our brain can fuse these images into an image that is free from under- and over-exposed problems, as shown in Fig. \ref{fig:bio}. 

Can we introduce this fusion mechanism of human visual system (HVS) to help build an accurate image enhancement algorithm?
Although many exposure fusion techniques have been proposed in the domain of HDR,
the additional images taken with different exposures are often not available for the low-light enhancement problem.
Fortunately, those images are highly correlated. 
The mapping function between two images that only differ in exposure
is called Brightness Transform Function (BTF).
Therefore, we can first use BTF to generate a series of multi-exposure images and then fuse those images to obtain the enhanced result.

\begin{figure*}[t]
\centering
\includegraphics[height=1.8in]{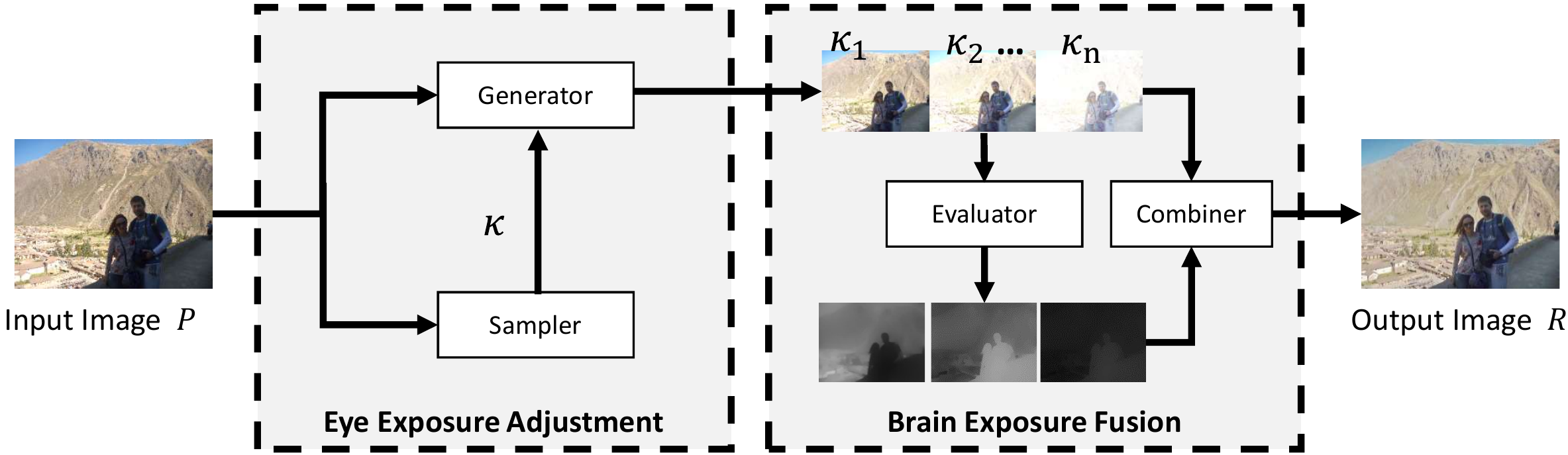}
\caption{
Our framework.
}
\label{fig:framework}
\end{figure*}

In this paper, we propose a multi-exposure fusion framework inspired by the HVS. 
There are two stages in our framework: Eye Exposure Adjustment and Eye Exposure Adjustment.
The first stage simulates the human eye to adjust the exposure, generating an multi-exposure image set.
The second stage simulates the human brain to fuse the generated images into the final enhanced result.
Based on our framework,
we propose a dual-exposure fusion method.
Specifically, we first employ the illumination estimation techniques to build the weight matrix for image fusion. Then we derives our camera response model based on observation. Next, we find the optimal exposure for our camera response model to generate the synthetic image that is well-exposed in the regions where the original image is under-exposed. Finally, we obtain the enhanced results by fusing the input image with the synthetic image using the weight matrix. 
Experiments on five challenging datasets are conducted to reveal the advantages of our method in comparison with other state-of-the-art methods.

\section{Related Work}

In general, image enhancement techniques can improve the subjective visual quality of input images and support the extraction of valuable information for some computer vision techniques \cite{ibrahim2007bpdhe,wang2013naturalness}. 
Low-light image enhancement, as one of enhancement techniques, can reveal the information of the under-exposed regions in an image.
Broadly speaking, existing low-light image enhancement techniques can be divided into two major categories: global enhancement and local enhancement.

\subsection{Global Enhancement Algorithms}

Global enhancement performs same processing on all image pixels regardless of their spatial distribution. 
Linear amplifying is a simple and straightforward global enhancement method. 
However, bright regions might be saturated after linear amplifying, causing some detail loss to the enhanced results.
To avoid the problem, some image enhancement methods adopt non-linear monotonic functions (e.g. power-law \cite{beghdadi1989contrast}, logarithm \cite{peli1990contrast} and gamma function \cite{gonzalez2008digital}) to perform enhancements. 
As another way to avoid saturation, histogram equalization (HE) \cite{jain1989fundamentals} can improve the contrast effectively and became a widely-used technique.
Many extensions of HE are proposed to take some restrictions into account such as brightness preservation 
\cite{ibrahim2007bpdhe,wang2005bpheme,chen2003mmbebhe} and contrast limitation \cite{reza2004clahe}. 
By extending the notion of the histogram,
some algorithms take spatial image features into consideration to further improve the performance \cite{lee2012ldr,celik2011cvc,arici2009wahe}. 
However, 
global enhancement may suffer from detail loss in some local areas because a global processing can not ensure all local areas be well enhanced.

%
\subsection{Local Enhancement Algorithms}
By making use of spatial information directly,
local enhancement can achieve better results and become the main-stream of recent techniques.
Local histogram equalization \cite{stark2000adaptive,abdullah2007dynamic}  
adopts the sliding window strategy to perform HE locally.
Based on the observation that the inverted low-light images are closed to hazy images, 
dehazing techniques are borrowed to solve low-light image enhancement in some methods \cite{dong2011fast,li2015low}. 
However, the basic models of above methods are lacking in physical explanation \cite{guo2017lime}. 
To provide a physical meaningful model for image enhancement,
Retinex theory assumes that the amount of light reaching observers 
can be decomposed into two parts: 
illumination and scene reflection. 
Most Retinex-based methods get enhanced results by removing the illumination part \cite{jobson1997properties,jobson1997multiscale,wang2014variational} while others \cite{fu2016weighted,wang2013naturalness,guo2017lime} keep a portion of the illumination to preserve naturalness.
Fu \etal \cite{fu2016mf} adjust the illumination components by fusing it with two enhanced illumination.
As far as we know,
there is no multi-exposure fusion method for this task
since lowlight enhancement problem usually takes a single image as input.

\section{Multi-Exposure Fusion Framework} 

Our framework mainly consists of four main components: 
the first component, named Multi-Exposure Sampler, 
determines how many images are required and the exposure ratio of each image to be fused;
the second component, named Multi-Exposure Generator, 
use a camera response model and the specified exposure ratio to synthetic multi-exposure images;
the third component, named Multi-Exposure Evaluator, 
determines the weight map of each image when fusing;
the last component, named Multi-Exposure Combiner, 
is to fuse the generated images to the final enhanced result based on the weight maps.
In this section, we introduce them one by one.






\subsection{Multi-Exposure Sampler}

Before we generate multi-exposure images, we need to determine how many images are required and their exposure ratios.
Since some images in the multi-exposure set cannot provide additional information,
taking these images into consideration is a waste of computation resources
and may even deteriorate the fused result.
A good sampler can use as few images as possible to reveal all the information in a scene by choosing appropriate exposure ratios.
The output of the sampler is a set of exposure ratios $\{ k_1, k_2, ... k_N\}$
where $N$ is the number of the images to be generated.

\subsection{Multi-Exposure Generator}

As aforementioned, images taken with different exposures are correlated.
Multi-Exposure Generator 
maps the input image into multi-exposure images according to the specified exposure ratio set.
The key part of the Multi-Exposure Generator is the camera response model used to find an appropriate BTF for mapping.
Given an exposure ratio $k_i$ and a BTF $g$, we can map the input image $\mathbf P$ to the $i$-th image in the
exposure set 
as
\begin{equation}
\mathbf P_i = g(\mathbf P, k_i).
\label{eqn:g}
\end{equation}

\subsection{Multi-Exposure Evaluator}%

To estimate the wellness of each pixel in the generated images. 
The Multi-Exposure Evaluator 
takes in an image and outputs a weight matrix that
indicates the wellness of each pixel. 
The weight matrix is nonuniform for all pixels: the well-exposed pixels are given a big weight while the poor-exposed pixels are given a small weight.
After all images are evaluated, the output matrices are pixel-wise normalized to ensure their summation equals one for each pixel as
\begin{equation}
\mathbf {\widetilde W_i} =  \mathbf W_i \oslash \sum_{i = 1}^N  \mathbf W_i,
\end{equation}
where $\oslash$ is the element-wise division, and $\mathbf W_i$ and $\mathbf {\widetilde W_i}$ are the $i$-th matrix before and after normalization, respectively.

\subsection{Multi-Exposure Combiner}

To obtain an image with all pixel well-exposed, we can simply fuse these images based on the weight matrix as
%
%
\begin{equation}
\mathbf R^c = \sum_{i = 1}^N  \mathbf {\widetilde W_i} \circ \mathbf P^c_i,
\label{eqn:J}
\end{equation}
where 
$c$ is the index of three color channels and $\mathbf R$ is the enhanced result.
Other fusion techniques like multi-scale fusion~\cite{mertens2009exposure} and
Boosting Laplacian Pyramid fusion~\cite{shen2014exposure}
can also be used to obtain a better fusion results.


\section{Dual-Exposure Fusion Algorithm} 

In this section, we use the proposed framework to design a low-light image enhancement algorithm.
To reduce complexity, we only generate one image with appropriate exposure and obtain the enhanced result by fusing the input image and the generated one.
Based on our framework, the fused image is defined as
\begin{equation}
\mathbf R^c = \mathbf {\hat{W}} \circ \mathbf P^c + ( \mathbf 1- \mathbf {\hat{W}} ) \circ g(\mathbf P^c, \hat{k}). 
\end{equation}
The enhancement problem can be divided into three parts: the determination of 
multi-exposure Evaluator ($\mathbf {\hat{W}}$),
multi-exposure generator ($g$), and multi-exposure sampler ($\hat{k}$). In the following subsections, we solve them one by one.

\subsection{Dual-Exposure Evaluator}

\begin{figure}[t]
\centering
\subfloat[Input image]{\includegraphics[width=0.7in]{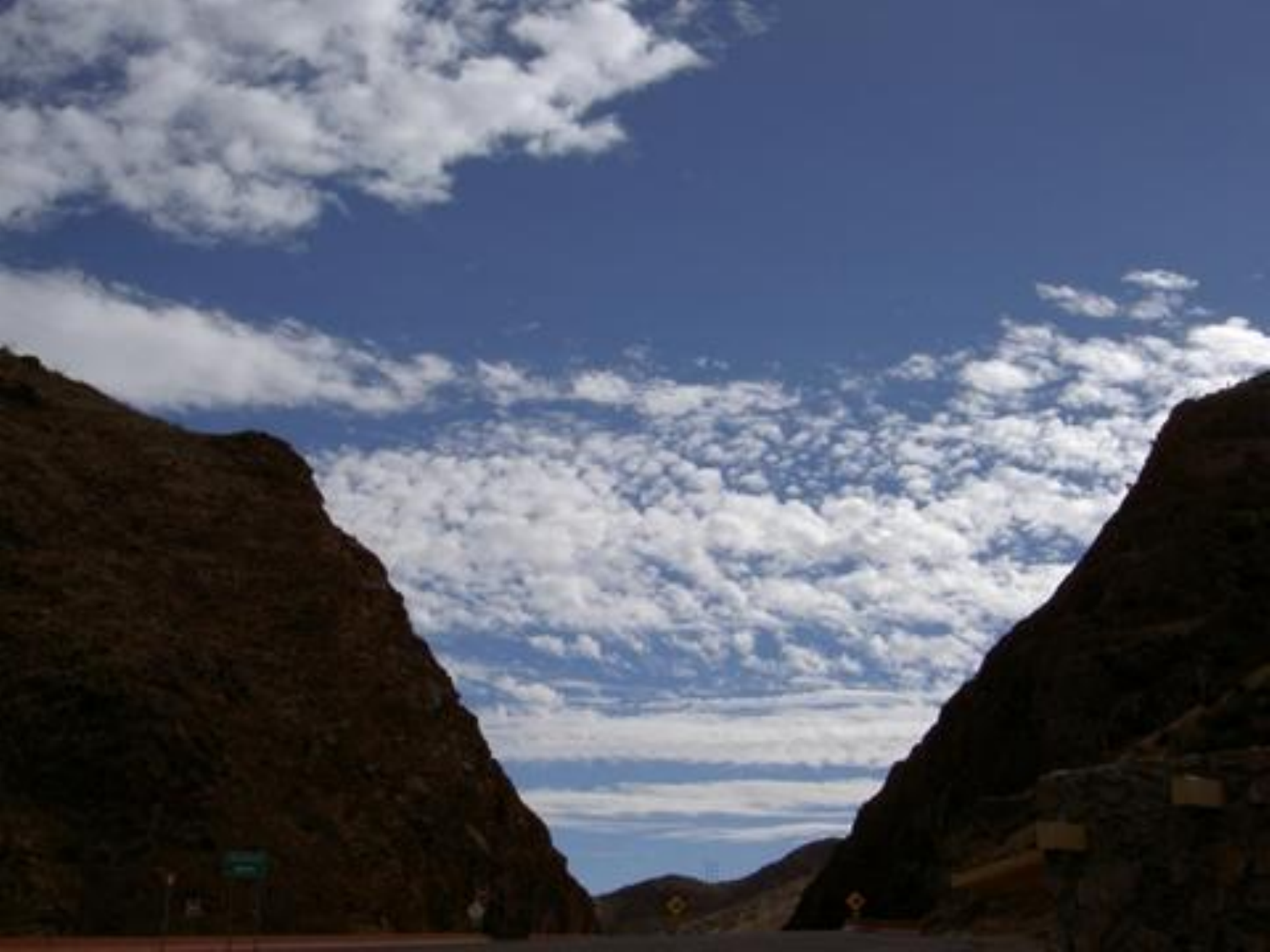}  }\hfil
\subfloat[$\mu$ = 0]{\includegraphics[width=0.7in]{mu_0.pdf}  }\hfil
\subfloat[$\mu$ = 0.5]{\includegraphics[width=0.7in]{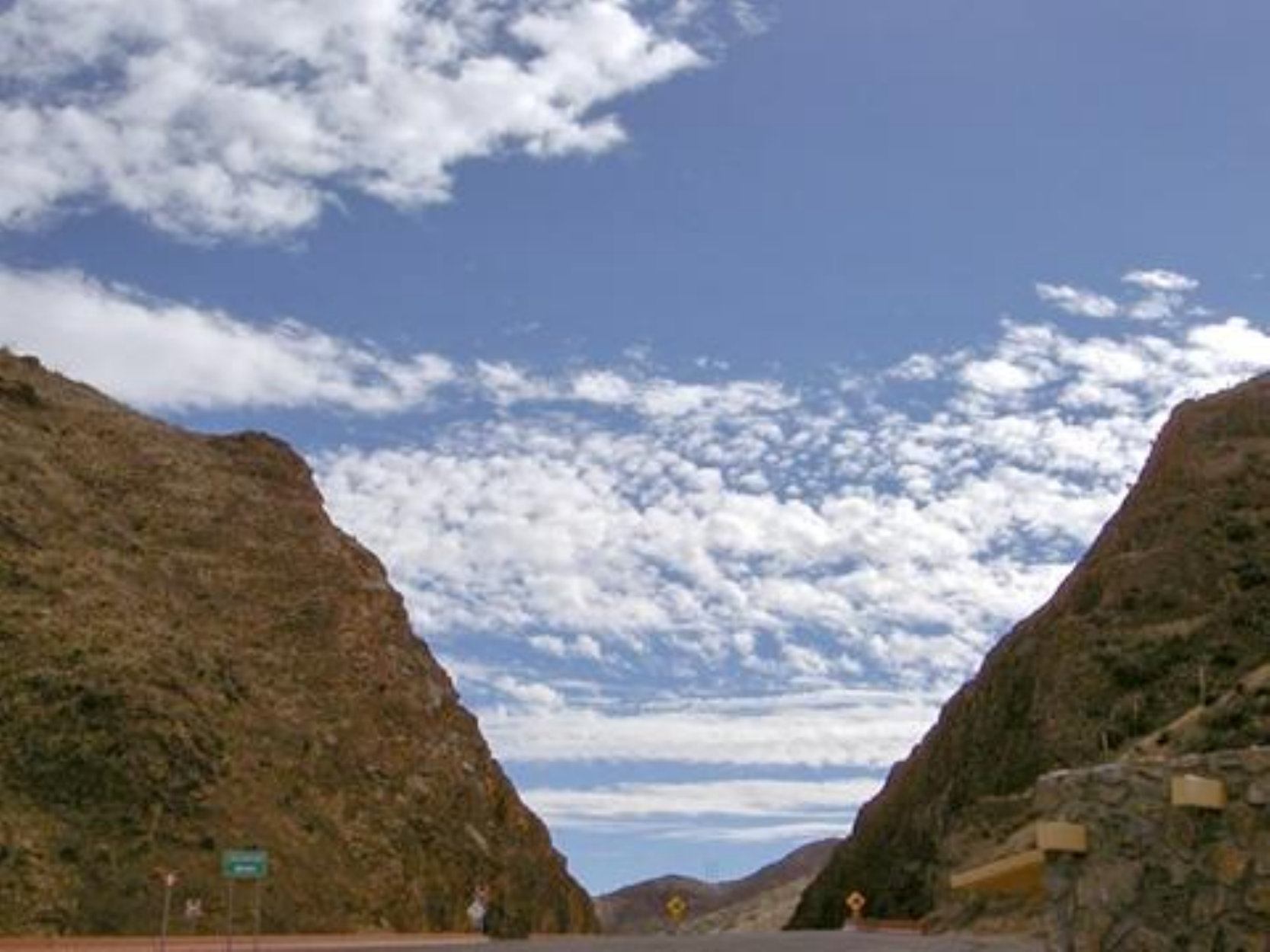} }\hfil
\subfloat[$\mu$ = 1]{\includegraphics[width=0.7in]{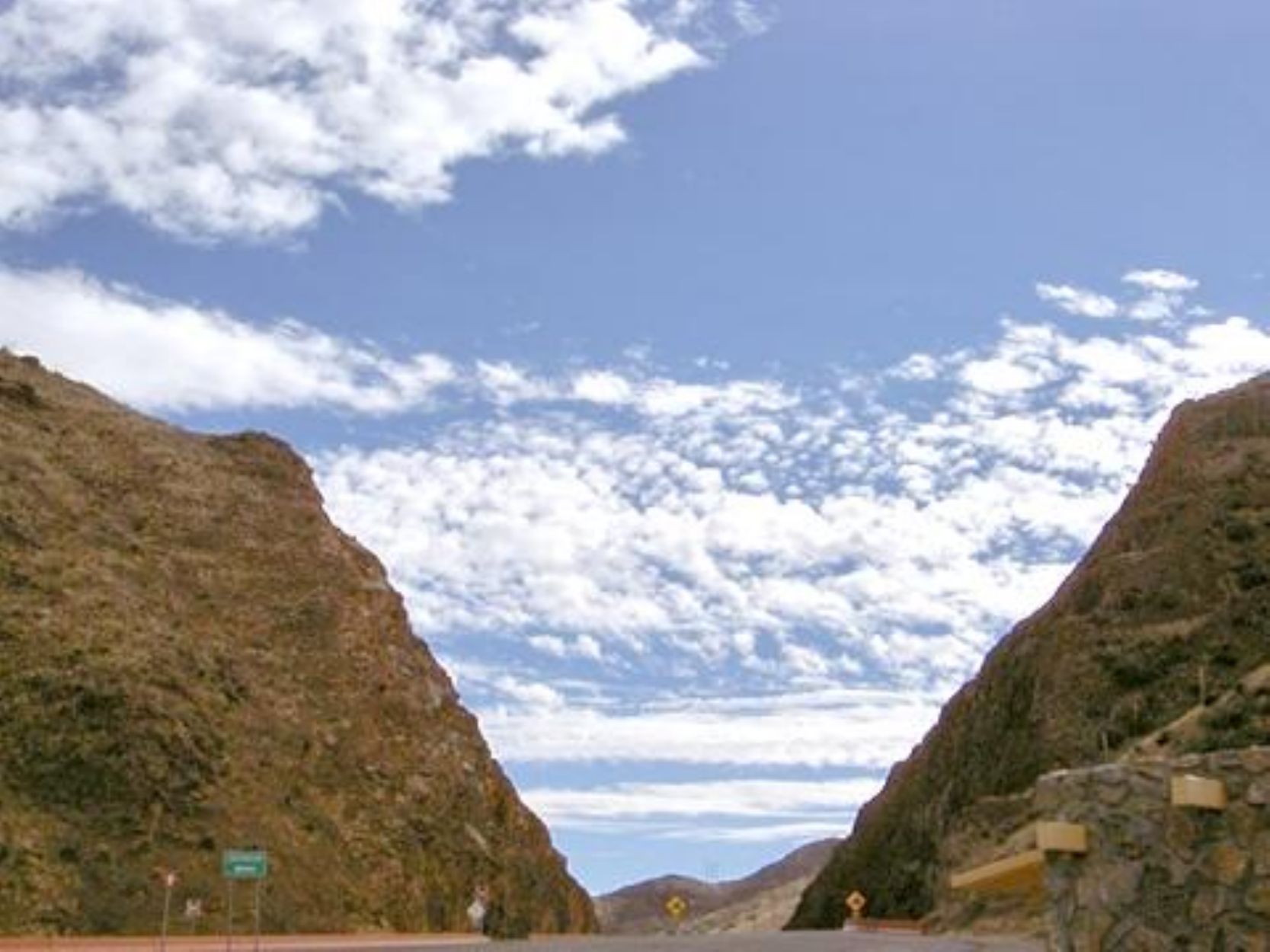}}\hfil
\\
\hspace{0.8in}
\subfloat[$\mu$ = 1.25]{\includegraphics[width=0.7in]{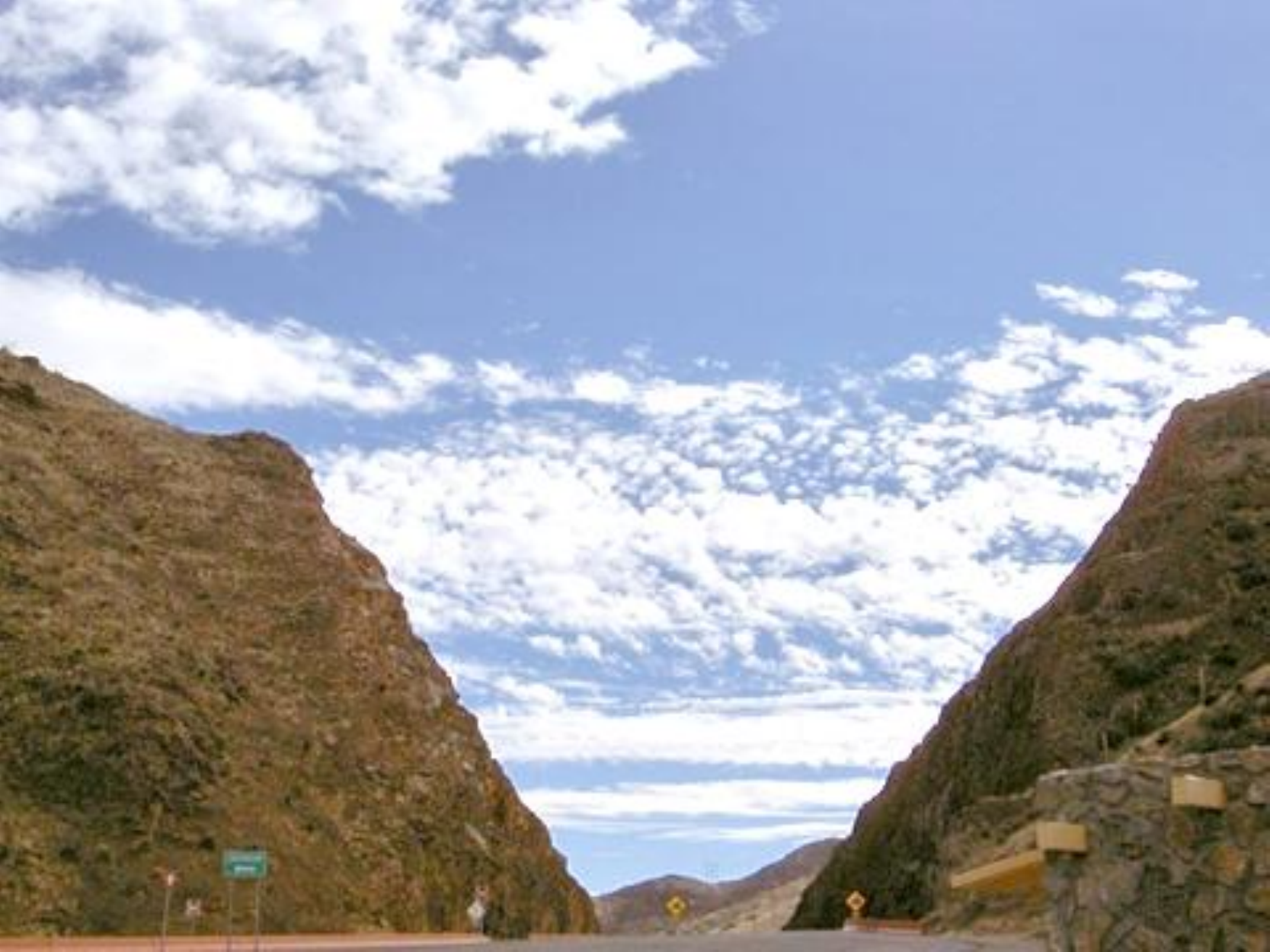}}\hfil
\subfloat[$\mu$ = 1.75]{\includegraphics[width=0.7in]{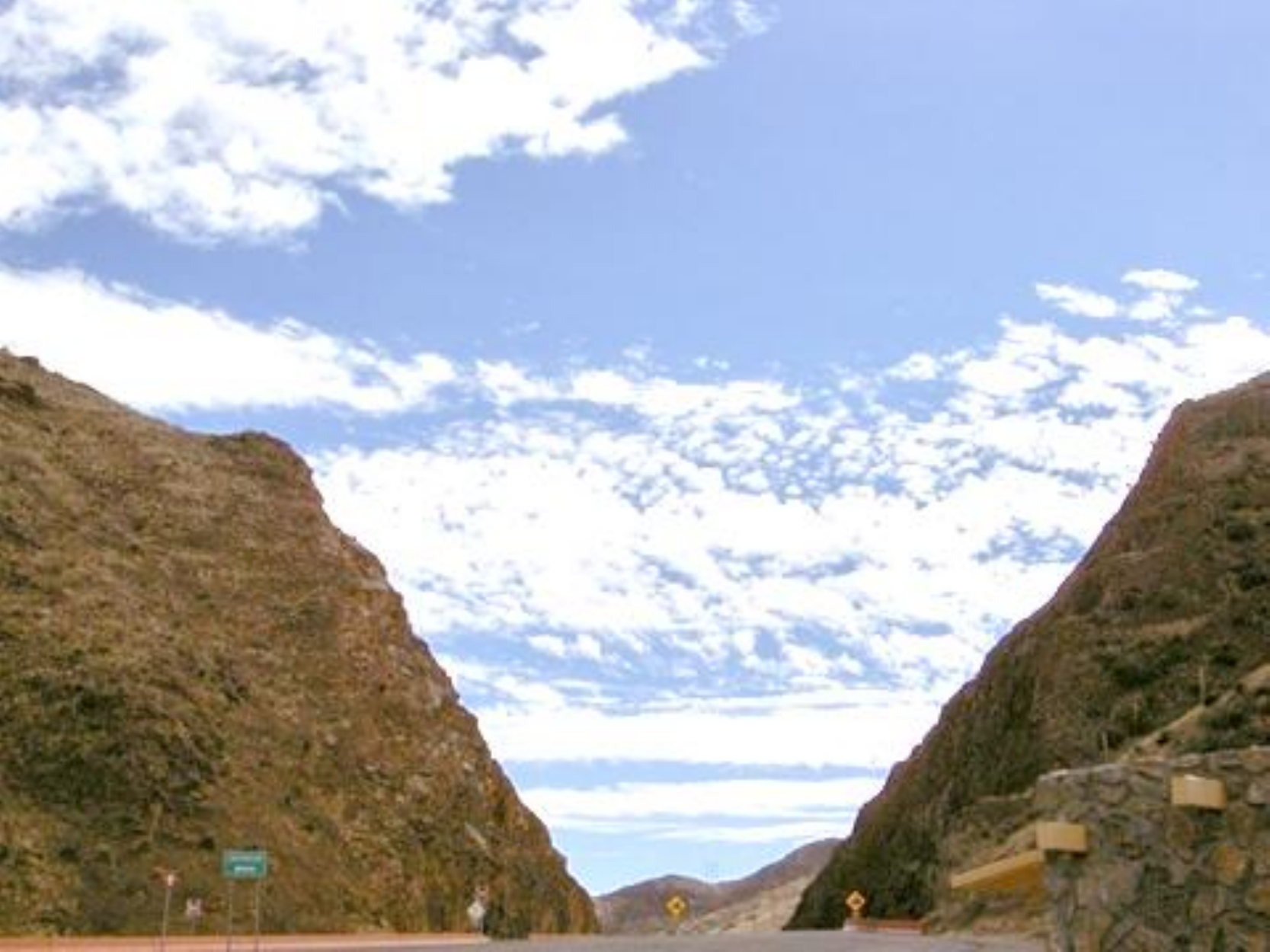}}\hfil
\subfloat[$\mu$ = 2.25]{\includegraphics[width=0.7in]{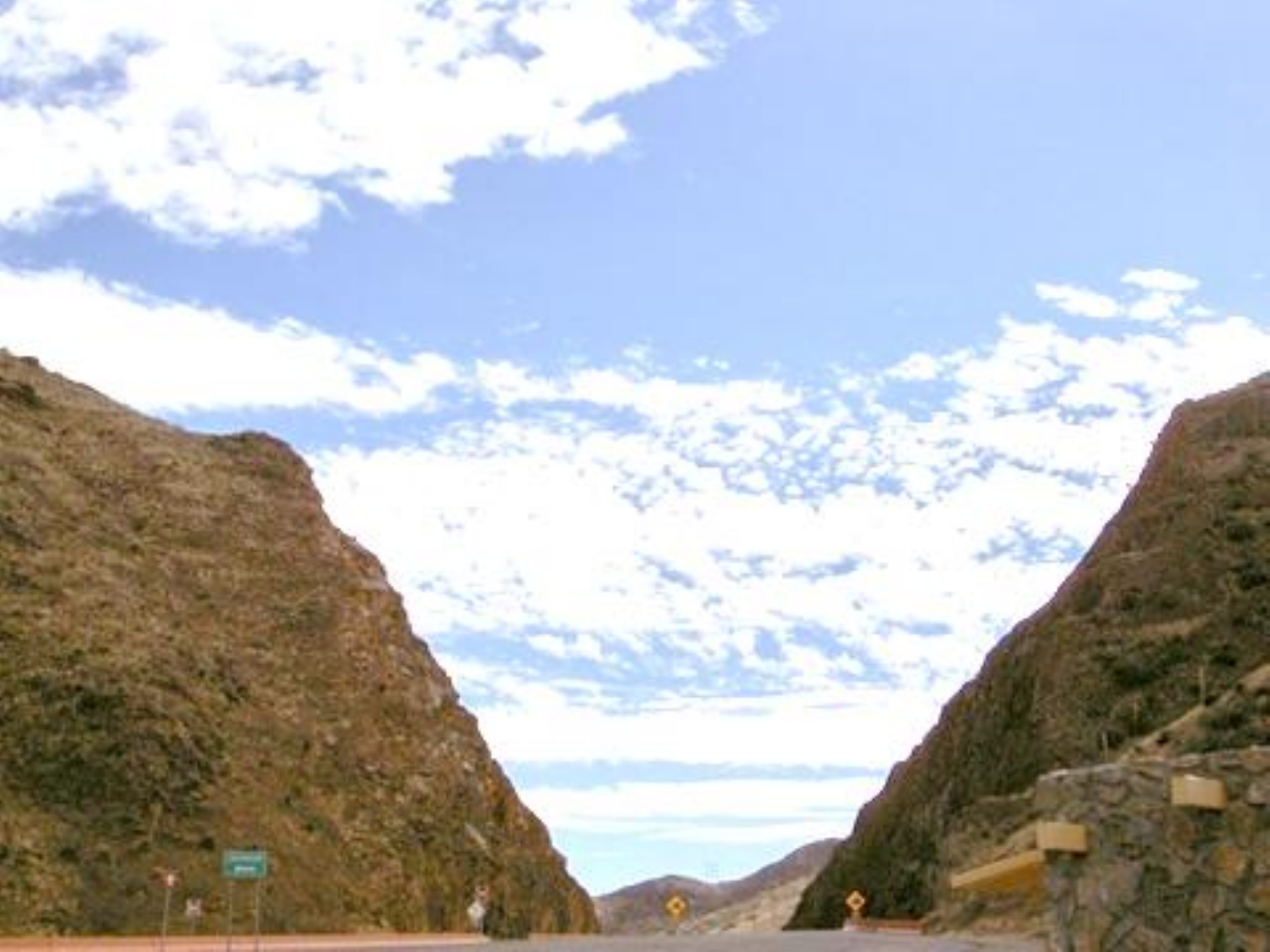}}
\caption{The enhanced results using different $\mu$.}
\label{fig_mu}
\end{figure}

The design of $\mathbf W$ is key to obtaining an enhancement algorithm that 
can enhance the low contrast of under-exposed regions while the contrast in well-exposed regions preserved.
We need to assign big weight values to well-exposed pixels and small weight values to under-exposed pixels.
Intuitively, the weight matrix is positively correlated with the scene illumination.
Since highly illuminated regions have big possibility of being well-exposed, 
they should be assign with big weight values to preserve their contrast.
We calculate the weight matrix as
\begin{equation}
\hat{\mathbf W} = \mathbf T ^ \mu
\end{equation}
where $\mathbf T$ is the scene illumination map and $\mu$ is a parameter controlling the enhance degree.
When $\mu = 0$, the resulting $\mathbf R$ is equal to $\mathbf P$, \ie, no enhancement is performed.
When $\mu = 1$, both the under-exposed pixels and  well-exposed pixels are enhanced.
When $\mu > 1$, pixels may get saturated and the resulting $\mathbf R$ suffers from detail loss.
As shown in Fig. \ref{fig_mu}.
In order to perform enhancement while preserve the well-exposed regions, we set $\mu$ to $0.5$.
The scene illumination map $\mathbf T$ is estimated by solving an optimization problem.

\subsubsection{Optimization Problem}
The lightness component can be used as an estimation of scene illumination.  
We adopt the lightness component as the initial estimation of illumination:
\begin{equation}
  \mathbf L(x) = \max\limits_{c \in \{R,G,B\} }\mathbf P_c(x) 
  \eqL{t0}
\end{equation}
for each individual pixel $x$. 
Ideal illumination should has local consistency for the regions with similar structures.
In other words, $\mathbf{T}$ should keep the meaningful structures of the image and remove the textural edges.
As in \cite{guo2017lime}, we refine $\mathbf{T}$ by solving the following optimization equation:
 \begin{equation}
  \min_{\mathbf{T}} ||\mathbf{T}-\mathbf L||^2_2+\lambda ||\mathbf{M} \circ \nabla{\mathbf{T}}||_1, 
  \eqL{optimization_problem}
\end{equation}
where $||*||_2$ and $||*||_1$ are the $\ell_2$ and $\ell_1$ norm, respectively. 
The first order derivative filter $\nabla$ contains $\nabla_h \mathbf{T}$ (horizontal) and $\nabla_v \mathbf{T}$ (vertical).
$\mathbf{M}$ is the weight matrix and $\lambda$ is the coefficient.
The first term of this equation is to minimize the difference between the initial map $\mathbf L$ and the refined map $\mathbf{T}$, 
while the second term maintains the smoothness of $\mathbf{T}$.

%


The design of $\mathbf{M}$ is important for the illumination map refinement.
A major edge in a local window contributes more similar-direction gradients than textures with complex patterns \cite{xu2012structure}. 
Therefore, The weight in a window that contains meaningful edges should be smaller than that in a window only containing textures. 
As a result, we design the weight matrix as 
\begin{equation}
\begin{aligned}
  \mathbf{M}_{d}(x) =\frac {1}{|\sum_{y \in \omega(x)}\nabla_d\mathbf L (y)|+\epsilon}, 
  \quad d \in \{ h, v\}, 
\end{aligned}
\eqL{weight}
\end{equation}
where $|*|$ is the absolute value operator,
$\omega(x)$ is the local window centered at the pixel $x$
and $\epsilon$ is a very small constant to avoid the zero denominator.


\subsubsection{Closed-Form Solution}
To reduce the complexity, 
we approximate \eqR{optimization_problem} as in \cite{guo2017lime}:
\begin{equation}
  \min_{\mathbf{T}} \sum_x \bigg( \big( \mathbf{T}(x)-\mathbf L (x) \big)^2 
  + \lambda \sum_{d \in \{ h, v \} } \frac {\mathbf{M}_{d}(x) \big( \nabla_d \mathbf{T}(x) \big)^2}{|\nabla_d \mathbf L (x)|+\epsilon } \bigg).
\end{equation}
As can be seen, the problem now only involves quadratic terms.
Let $\mathbf m_d$, $\mathbf l$, $\mathbf{t}$ and $\nabla_d \mathbf l$ denote the vectorized version of $\mathbf M_d$, $\mathbf L$, $\mathbf{T}$ and $\nabla_d \mathbf L$ respectively.
Then
the solution can be directly obtained by solving the following linear function.
\begin{equation}
  (\mathbf{I}+\lambda \sum_{d \in \{ h, v \} } (\mathbf{D_d^\intercal}Diag( {\mathbf{m}_d} \oslash ({|\nabla_d \mathbf l| + \epsilon}))\mathbf{D_d})\mathbf{t}= \mathbf l
  \eqL{linear_eq}
\end{equation}
where $\oslash$ is the element-wise division, $\mathbf{I}$ is the unit matrix, the operator $Diag(\mathbf v)$ is to construct a diagonal matrix using vector $\mathbf v$, 
and $\mathbf{D_d}$ are the Toeplitz matrices from the discrete gradient operators with forward difference.

\begin{figure}[t]
\centering
\subfloat[]{\includegraphics[width=1in]{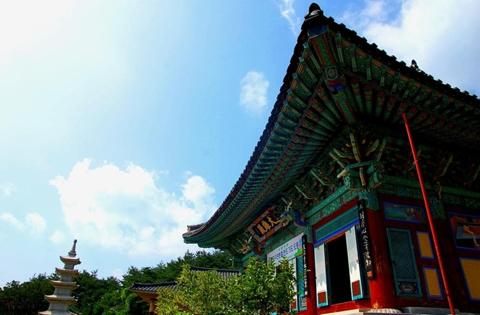}}
\subfloat[]{\includegraphics[width=1in]{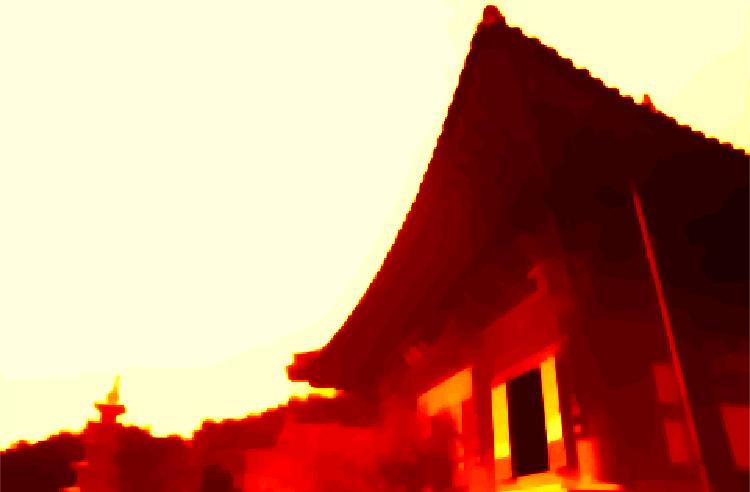}}
\subfloat[]{\includegraphics[width=1in]{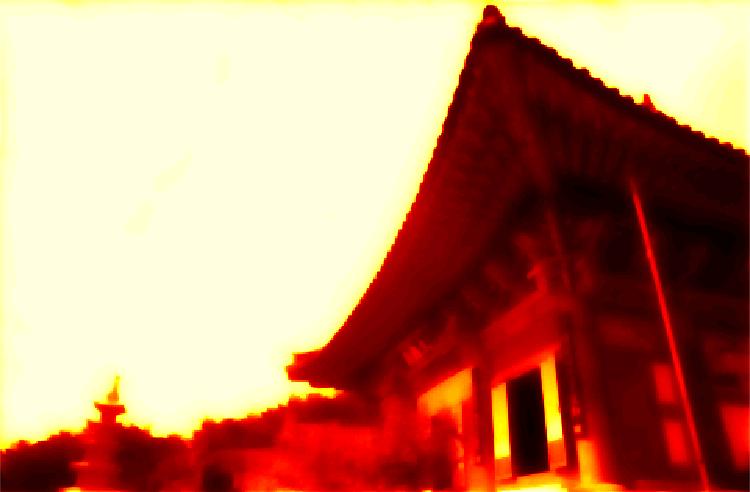}}
\\ 
\hspace{1in}
\subfloat[]{\includegraphics[width=1in]{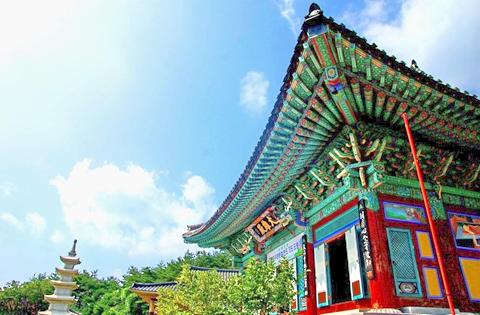}}
\subfloat[]{\includegraphics[width=1in]{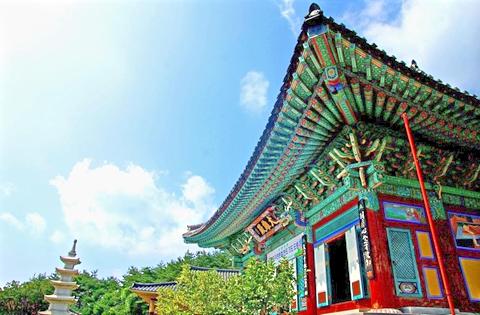}}
\caption{
   (a) Input image.
   (b) Estimated illumination map by \cite{guo2017lime} (0.21s).
   (c) Our illumination map (0.15s). 
   (d) Enhanced result using (b).
   (e) Enhanced result using (c). 
}
\label{fig:map}
\end{figure}

The main difference between our illumination map estimation method and that in \cite{guo2017lime}
is the design of weight matrix $\mathbf M$. 
We adopt a simplified strategy which can yield similar results as in \cite{guo2017lime}.
As shown in Fig. \ref{fig:map}.
although the illumination map in \cite{guo2017lime} is sharper than ours, our method is more time-saving while the two enhanced results show no significant visual difference.

\subsection{Dual-Exposure Generator}  
In this section, we present a camera response model to implement the Multi-Exposure Generator. 
A camera response model consists of two parts: Camera Response Function (CRF) model and BTF model.
The parameters of CRF model is determined only by camera while 
that of BTF model is determined by camera and exposure ratio.
In this subsection, we first propose our BTF model based on the observation of two different exposure images. 
Then we derive the corresponding CRF model by solving the comparametric equation.
Finally, we discuss how to determine the model parameters and present the final form of $g$.
\subsubsection{BTF Estimation} 

\begin{figure}[b]
\centering
\includegraphics[width=2.7in]{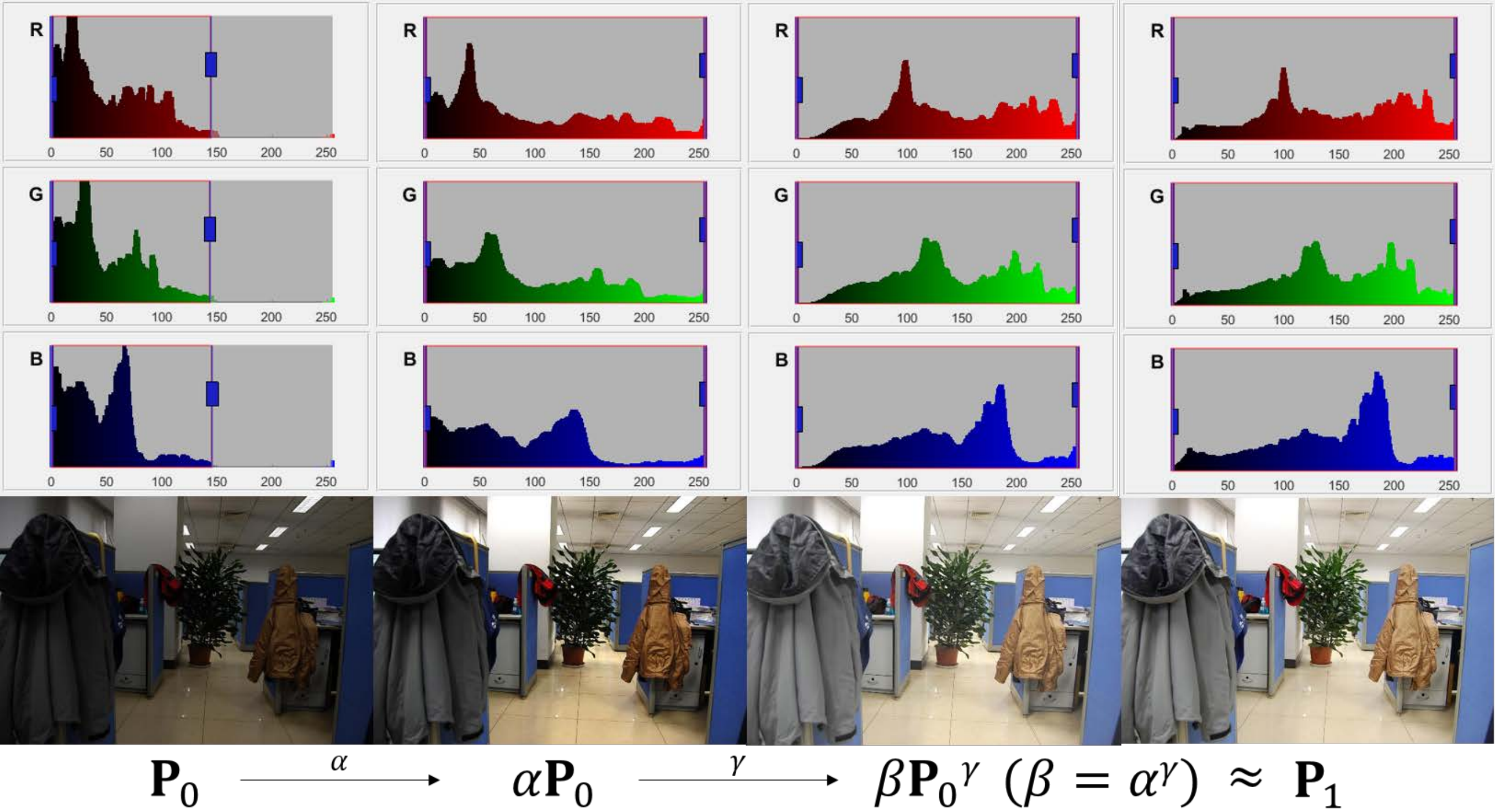}
\caption{Observation. From left to right: An under-exposure image $\mathbf{P}_0$, apply multiplication $\alpha \mathbf{P}_0$, apply gamma function $(\alpha \mathbf{P}_0)^\gamma$ and the well-exposure image $\mathbf{P}_1$ under the same scene. The histograms of red, green and blue color channels are plotted above the corresponding image respectively.}
\label{fig:observation_hist}
\end{figure}




To estimate the BTF $g$,
we select a pair of images $\mathbf{P}_0$ and $\mathbf{P}_1$ that differ only in exposure.
Then we plot their histograms of each color channel, as shown in \figR{observation_hist}.
Noticing that the histograms of the under-exposed image mainly concentrate in low-brightness area, 
if we perform linear amplification of pixel values before traditional gamma correction,
then the resulting image will be very close to the real well-exposed image. 
Therefore, we can use a two-parameter function to describe the BTF model as
\begin{equation}
 \mathbf{P}_1 = g(\mathbf{P}_0,k) = \beta \mathbf{P}_0 ^ \gamma,
 \label{eqn_g}
\end{equation}
where $\beta$ and $\gamma$ are parameters in our BTF model related to exposure ratio $k$.
The observation also shows
that different color channels have approximately same model parameters.
The underlying reason is that the response curves of different color channels 
are approximately identical for general cameras.




\subsubsection{CRF Estimation}



In our BTF model, $\beta$ and $\gamma$ are determined by the camera parameters and exposure ratio $k$.
To find their relationship, we need to obtain the corresponding CRF model.
The CRF model can be derived by solving the following comparametric equation (plug $g = \beta f^ \gamma$ to $f(kE) = g(f(E))$:
\begin{equation}
\eqL{our_comparametric}
f(kE) = \beta f(E)^\gamma.
\end{equation}
The closed-form solution of $f$ is provided in as follows (see Appendix for detail):
\begin{equation}
  f(E) = 
  \begin{cases}
    e^{b(1 - E^a) }, & \text{if } \gamma \neq 1, \\
    E^c, & \text{if } \gamma = 1. 
  \end{cases}
\eqL{fin}
\end{equation}
where $a$ and $b$ are model parameters in the case of $\gamma \neq 1$:
\begin{equation}
  a = \log_{k}{\gamma}, \quad b =\frac{\ln \beta}{1-\gamma};
\eqL{a_and_b}
\end{equation}
And $c$ is a model parameter in the case of $\gamma = 1$:
\begin{equation}
  c = \log_k \beta.
\eqL{c}
\end{equation}

\begin{figure*}[t]
\centering
\includegraphics[width=6in]{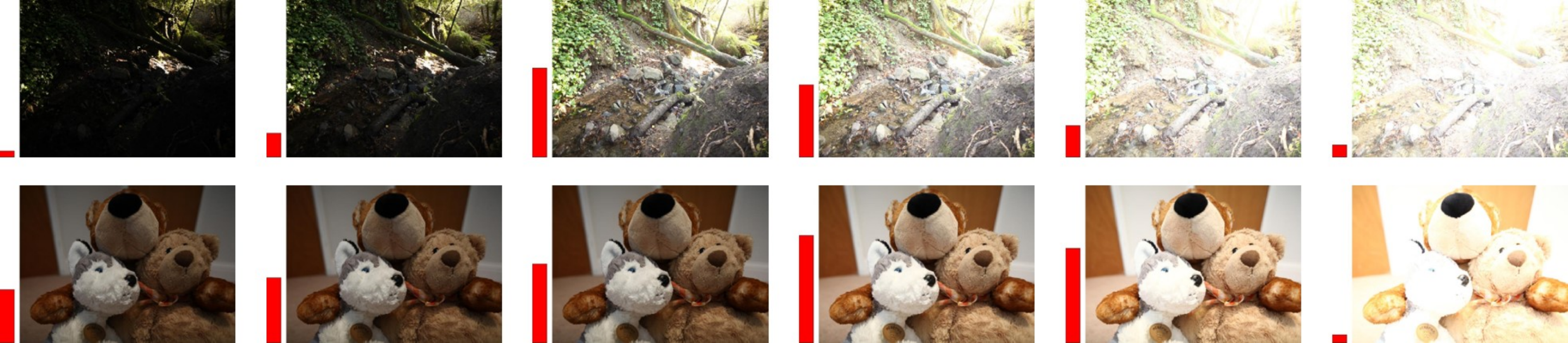}
\caption{
Examples of multi-exposure image sets and their entropy values. 
The red bar on the left of each image shows the exponential of its entropy value.
}
\label{fig:entropy}
\end{figure*}

Two CRF models can be derived from two cases of \eqR{fin}. 
When $\gamma =1$, the CRF model is a power function and the BTF model is a simple linear function.
As some camera manufacturers design $f$ to be a gamma curve, it can fit these cameras perfectly.
When $\gamma \neq 1$, the CRF model is a two-parameter function and the BTF model is a non-linear function.
Since the BTF is non-linear for most cameras, we mainly consider the case of $\gamma \neq 1$.
Our BTF $g$ is solved as
\begin{equation}
  g({\mathbf P}, k) = e^{b(1 - k^a)  }  {\mathbf P}^{(k^a)}.
\end{equation}
where $\beta$ and $\gamma$ are two model parameters that can be calculated from
camera parameters $a$, $b$ and exposure ratio $k$.
We assume that no information about the camera is provided and use a fixed camera parameters 
($a = -0.3293, b = 1.1258$) that can fit most cameras.




\subsection{Dual-Exposure Sampler} 


As aforementioned, our algorithm only generate one image.
So, in this subsection, we only need to determinate the optimal exposure ratio of the generated image.
In order to represent as many information as possible using only the input image and the generated one, we find the best exposure ratio so that the synthesis image
is well-exposed in the regions where the original image under-exposed.

First, we exclude the well-exposed pixels and obtain an image that is globally under-exposed.
We simply extract the low illuminated pixels as 
\begin{equation} 
  \mathbf{Q} = \{ \mathbf P(x) | \mathbf{T}(x) < 0.5 \},
\end{equation}
where $\mathbf{Q}$ contains only the under-exposed pixels.




The brightness of the images under different exposures changes significantly while the color is basically the same. 
Therefore, 
we only consider the brightness component when estimating $k$.
The brightness component $\mathbf{B}$ is defined as the geometric mean of three channel:
%
\begin{equation}
  \mathbf{B} \coloneqq \sqrt[3]{\mathbf{Q}_r \circ \mathbf{Q}_g \circ \mathbf{Q}_b},
\label{eqn_Y}
\end{equation}
where $\mathbf{Q}_r$, $\mathbf{Q}_g$ and $\mathbf{Q}_b$ are the red, green and blue channel of the input image $\mathbf{Q}$ respectively.
We use the geometric mean instead of other definitions (\eg arithmetic mean and weighted arithmetic mean) since it has the same BTF model parameters ($\beta$ and $\gamma$) with all three color channels, as shown in \eqR{samebeta}.
\begin{equation}
\eqL{samebeta}
 \begin{aligned}
 \mathbf B^\prime & \coloneqq \sqrt[3]{\mathbf{Q}^\prime_r \circ \mathbf{Q}^\prime_g  \circ \mathbf{Q}^\prime_b } \\
 & = \sqrt[3]{(\beta \mathbf{Q}_r ^\gamma) \circ (\beta \mathbf{Q}_g ^\gamma) \circ (\beta \mathbf{Q}_b ^\gamma)} 
  = \beta (\sqrt[3]{\mathbf{Q}_r \circ \mathbf{Q}_g \circ \mathbf{Q}_b})^\gamma
 \\ & = \beta \mathbf B^\gamma.
  \end{aligned}
\end{equation}


The visibility of a well-exposed image is higher than that of an under/over-exposed image and it can provide a richer information for human.
Thus, the optimal $k$ should provide the largest amount of information. 
To measure the amount of information, we employ the image entropy which is defined as
\begin{equation}
  \mathcal{H}(\mathbf{B}) = - \sum_{i=1}^{N} p_i \cdot \log_2 p_i,
\end{equation}
where $p_i$ is the $i$-th bin of the histogram of $\mathbf{B}$ which counts the number of data valued in $[\frac{i}{N}, \frac{i+1}{N})$
and $N$ is the number of bins ($N$ is often set to be 256).
As shown in Fig. \ref{fig:entropy}, the image entropy of a well-exposed image is higher than that of an under/over-exposed image. Therefore, it is reasonable to use the entropy to find the optimal exposure ratio.
%
%
The optimal exposure ratio $\hat{k}$ is calculated by maximizing the image entropy of the enhancement brightness as
\begin{equation} 
  \hat{k} = \argmax_{k} \mathcal{H}( g( \mathbf{B}, k ) ). 
\label{eqn_optimal_k}
\end{equation}
Since the image entropy increases first and then decreases with the increase of the exposure ratio, $\hat{k}$ can be solved by one-dimensional minimizer. 
%
To improve the calculation efficiency, we resize the input image to $50 \times 50$ when optimizing $k$. 

\begin{figure*} [t]
\centering
\includegraphics[width=7in]{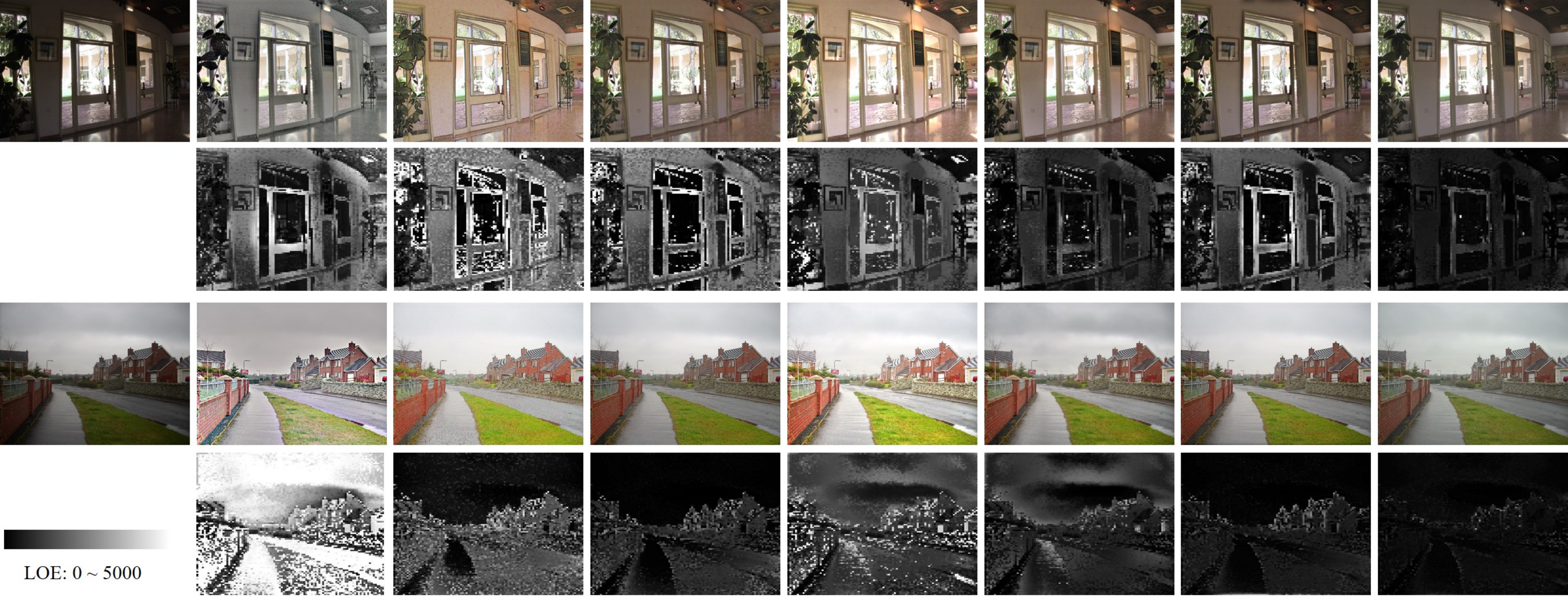}
\\
\small
\qquad 
Input \hfill MSRCR \hfill Dong \hfill NPE \hfill LIME  \hfill MF \hfill SRIE \hfill Ours
\qquad 
\figLC{lightnessDistortion}{
Comparison of lightness distortion.
The odd rows show the original image and the results of various enhancement methods, 
and the even rows show the visualization of each method's lightness distortion ($RD$).}
\end{figure*}


\begin{table*}[t]
\centering
\caption{Quantitative measurement results of lightness distortion (LOE).}
\label{tbl:loe}
\begin{tabular}{@{}rrrrrrrrr@{}}
\toprule
              & VV              & LIME-data       & NPE-data        & NPE-ex1         & NPE-ex2         & NPE-ex3         & DICM            & MEF             \\ \midrule
MSRCR         & 2727.7          & 1835.5          & 1889.7          & 1870.3          & 1944.7          & 1776.3          & 1795.3          & 1686.2          \\
Dong          & 853.35          & 1244            & 1012            & 1426.1          & 1096.3          & 1466.2          & 1180            & 1065.4          \\
NPE           & 820.93          & 1471.3          & 646.34          & 840.83          & 775.82          & 1130            & 662.29          & 1158.2          \\
LIME          & 1274.7          & 1323.8          & 1119.6          & 1321.9          & 1215.4          & 1319.1          & 1260.8          & 1079.4          \\
MF            & 470.93          & 629.82          & 488.07          & 851.87          & 541.85          & 749.72          & 667.45          & 525.95          \\
SRIE          & 551.39          & 823.61          & 533.24          & 653.05          & 564.49          & 760.76          & 623.32          & 754.2           \\
\textbf{Ours} & \textbf{287.22} & \textbf{478.57} & \textbf{308.12} & \textbf{319.93} & \textbf{323.72} & \textbf{378.65} & \textbf{351.82} & \textbf{325.86} \\
\bottomrule
\end{tabular}
\end{table*}

\section{Experiments} \label{sec-exp}


To evaluate the performance of our method,
we compare it with several state-of-the-art methods,
including 
Multi Scale Retinex with Color Restoration (MSRCR)~\cite{petro2014multiscale},
Naturalness Preserved Enhancement algorithm (NPE)~\cite{wang2013naturalness},
dehazing based method (Dong)~\cite{dong2011fast},
Multi-deviation Fusion method (MF)~\cite{fu2016mf},
Illumination Estimation based method (LIME) \cite{guo2017lime}
and
Simultaneous Reflection and Illumination Estimation (SRIE)~\cite{fu2016srie}.
We test those methods on hundreds of low-light images from five public datasets:
VV~\cite{vonikakis2017evaluation}, 
LIME-data \cite{guo2017lime}, 
NPE-data and its extension (NPE, NPE-ex1, NPE-ex2 and NPE-ex3)~\cite{wang2013naturalness},
DICM~\cite{lee2012ldr},
and
MEF \cite{ma2015perceptual}.
The datasets are briefly introduced as follows:

\textbf{VV}\footnote{\url{https://sites.google.com/site/vonikakis/datasets}}. 
This dataset is collected by Vassilios Vonikakis in his daily life to provide
the most challenging cases for enhancement. 
Each image in the dataset has a part that is correctly exposed and a part that is severely under/over-exposed.
A good enhancement algorithm should enhance the under/over-exposed regions while not affect the correctly exposed one.

\textbf{LIME-data}\footnote{\url{http://cs.tju.edu.cn/orgs/vision/~xguo/LIME.htm}}. 
This dataset contains 10 low-light images used in \cite{guo2017lime}.

\textbf{NPE}\footnote{\url{http://blog.sina.com.cn/s/blog_a0a06f190101cvon.html}}. 
This dataset contains 85 low-light images downloaded from Internet.
NPE-data contains 8 outdoor nature scene images which are used in \cite{wang2013naturalness}.
NPE-ex1, NPE-ex2 and NPE-ex3 are three supplementary datasets including cloudy daytime, daybreak, nightfall and nighttime scenes.

\textbf{DICM}\footnote{\url{http://mcl.korea.ac.kr/projects/LDR/LDR_TEST_IMAGES_DICM.zip}}. It contains 69 captured images from commercial digital cameras collected by \cite{lee2012ldr}.

\textbf{MEF}\footnote{\url{https://ece.uwaterloo.ca/~k29ma/}}. 
This dataset was provided by \cite{ma2015perceptual}. 
It contains 17 high-quality image sequences including natural sceneries, indoor and outdoor views and man-made architectures.
Each image sequence has several multi-exposure images, we select one of poor-exposed images as input to perform evaluation.





In order to maintain the fairness of the comparison, 
all the codes are in Matlab and all the experiments are conducted on a PC running Windows 10 OS with 64G RAM and 3.4GHz CPU (GPU acceleration is not used).
The parameters of our enhancement algorithm are fixed in all experiments:
$\lambda = 1$, $\epsilon = 0.001$, $\mu = 1/2$, and the size of local window $\omega(x)$ is 5. 
The most time-consuming part of our algorithm is illumination map optimization.
We employ the multi-resolution preconditioned conjugate gradient solver ($\mathcal{O}(N)$) to solve it efficiently. 

\begin{figure*} [t]
\centering
\includegraphics[width=7in]{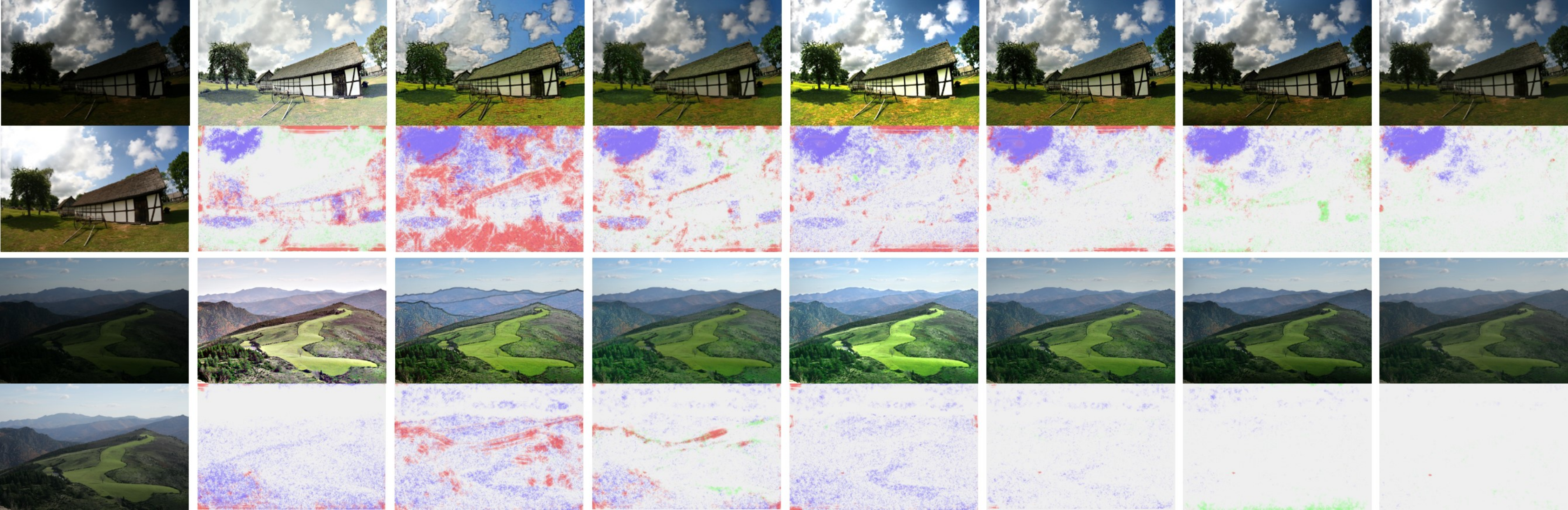}
\\
\small
\qquad 
Input \hfill MSRCR \hfill Dong \hfill NPE \hfill LIME  \hfill MF \hfill SRIE \hfill Ours
\qquad \qquad \qquad
\\
\caption{
Comparison of contrast distortion. 
The loss of visible contrast is marked in green, the amplification of invisible contrast is marked in blue, and the reversal of visible contrast is marked in red. Different shades of color represent different
degrees of distortion.
}
\label{fig:resultHdr}
\end{figure*}

\begin{table*}[]
\centering
\caption{Quantitative measurement results of visual information fidelity (VIF).}
\label{tbl:vif}
\begin{tabular}{@{}rrrrrrrrr@{}}
\toprule
              & VV               & LIME-data        & NPE-data         & NPE-ex1          & NPE-ex2          & NPE-ex3          & DICM             & MEF              \\ \midrule
MSRCR         & 0.42134          & 0.24045          & 0.41425          & 0.29822          & 0.38625          & 0.65951          & 0.44966          & 0.27995          \\
Dong          & 0.50477          & 0.32519          & 0.43440          & 0.38049          & 0.41687          & 0.50236          & 0.52637          & 0.35322          \\
NPE           & 0.69006          & 0.50885          & 0.71471          & 0.58572          & 0.67769          & 0.74368          & 0.72497          & 0.52376          \\
LIME          & 0.34932          & 0.20500          & 0.33934          & 0.28473          & 0.29498          & 0.38248          & 0.41498          & 0.22764          \\
MF            & 0.72414          & 0.44752          & 0.63859          & 0.57687          & 0.61976          & 0.71747          & 0.70703          & 0.51293          \\
SRIE          & 0.65968          & 0.52139          & 0.66528          & 0.58634          & 0.63547          & 0.69676          & 0.67866          & 0.55311          \\
\textbf{Ours} & \textbf{0.76098} & \textbf{0.74205} & \textbf{1.04930} & \textbf{0.70719} & \textbf{0.69787} & \textbf{0.76016} & \textbf{0.74524} & \textbf{0.60063} \\
\bottomrule
\end{tabular}
\end{table*}

\begin{figure*} 
\centering
\includegraphics[width=7in]{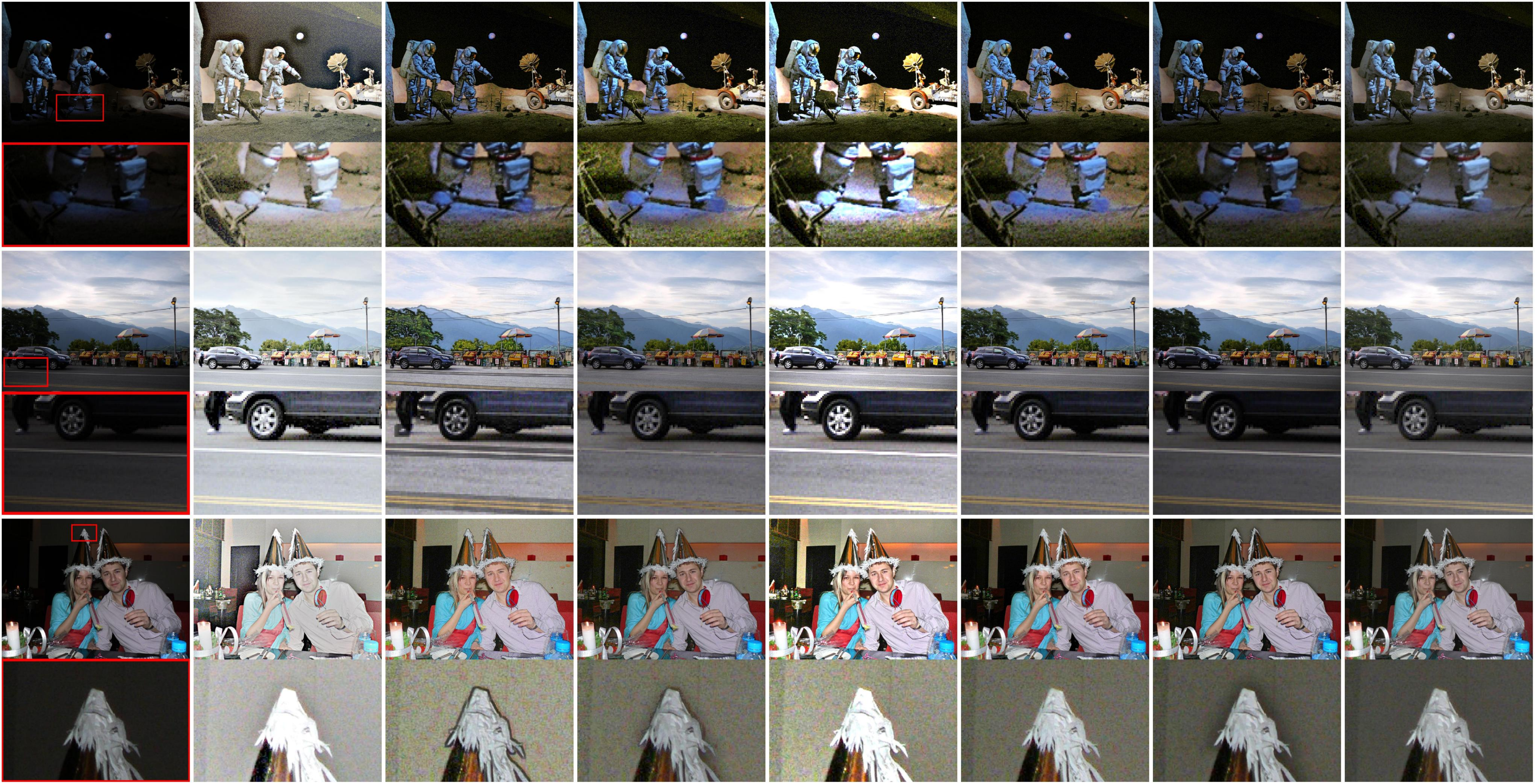}
\\
\small
\qquad 
Input \hfill MSRCR \hfill Dong \hfill NPE \hfill LIME  \hfill MF \hfill SRIE \hfill Ours
\qquad \qquad \qquad\\
\caption{Visual comparison among the competitors on different scenes.} 
\label{fig:more}
\end{figure*}

\begin{figure} [b]
\centering
\includegraphics[width=2in]{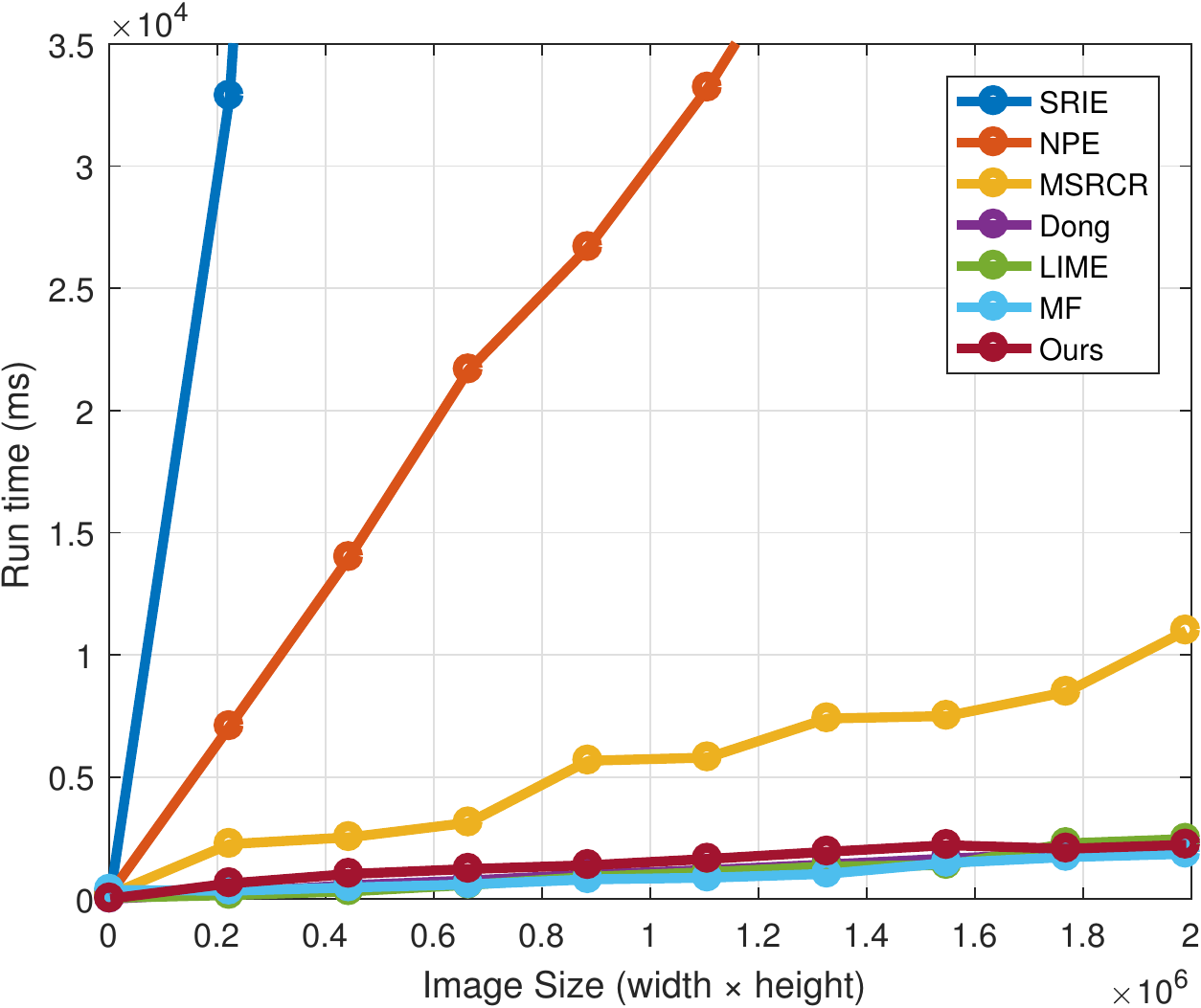}
\caption{
Time comparison among different methods with varying image sizes
} 
\label{fig:time}
\end{figure} 

\subsection{Lightness Distortion}

We use lightness order error (LOE) to objectively measure the lightness distortion of enhanced results. 
LOE is defined as 
\begin{equation}
LOE = \frac{1}{m} \sum_{x=1}^m RD(x)
\eqL{loe}
\end{equation}
where $RD (x)$ is the relative order difference of the lightness between the original image $P$ and its enhanced
version $P^\prime$ for pixel $x$, which is defined as follows:
\begin{equation}
RD (x) = \sum_{y=1}^m U\Big( \mathbf{L}(x), \mathbf{L}(y) \Big) \oplus U\Big( \mathbf{L}^\prime(x), \mathbf{L}^\prime(y) \Big),
\end{equation}
where m is the pixel number, $\oplus$ stands for the exclusive-or operator, 
$\mathbf{L}(x)$ and $\mathbf{L}^\prime(x)$ are the lightness component at location $x$ of the input images and the enhanced images, respectively.
The function $U(p, q)$ returns 1 if $p >= q$, 0 otherwise. 

As suggested in \cite{guo2017lime,wang2013naturalness}, down-sampling is used to reduce the complexity of computing LOE. 
We notice that LOE may change significantly when an image is down-sampled to different sizes
since $RD$ will increase as the pixel number $m$ increases.
Therefore, we down-sample all images to a fixed size. Specifically, we collect 100 rows and columns evenly to form a $100 \times 100$ down-sampled image.

As shown in Table \ref{tbl:loe},
our algorithm outperforms the others in all datasets.
This means that our algorithm can maintain the naturalness of images well.
We also provide a visualization of lightness distortion on two cases in \figR{lightnessDistortion},
from which,
we can find our results have the smallest lightness distortion.
The results of MSRCR lose the global lightness order and suffer from severe lightness distortion.
Although the results of LIME is visually pleasant, they are full of lightness distortion.
The results of Dong, NPE, MF and SRIE can only retain the lightness order in the well-exposed regions. 




\subsection{Contrast Distortion}
As aforementioned, the image that only differ in exposures can be used as a reference for evaluating 
the accuracy of enhanced results. 
DRIM (Dynamic Range Independent Metric) \cite{aydin2008drim} 
can measure the distortion of image contrast without the interference
of change in image brightness. 
We use it to visualize the contrast difference between the
enhanced result and the reference image.
As shown in Fig. \ref{fig:resultHdr}, the proposed method obtains the most realistic results with the least distortion. 





\subsection{Visual Information Distortion}

To measure the distortion of visual information, we employ
Visual Information Fidelity (VIF)~\cite{sheikh2006image} in reverse mode.
As a full reference image quality assessment index,
VIF models the quality assessment problem as a information fidelity criterion that quantifies the mutual information
between the reference image $C$ and the distorted image $F$
relative to the information of $C$ extracted by the HVS.
VIF measure is given by
\begin{equation}
	VIF = \frac{I(C;F)}{I(C;E)},
\end{equation}
where $E$ is the image that the HVS perceives. 
The mutual information $I(C;F)$ and $I(C;E)$ represent the information that could be extracted by the brain in the reference and the test images respectively.

Like most of full reference image quality assessment methods, 
VIF were designed for and tested on degraded images.
The normal version of VIF
treats the original image as the reference image and 
the image outputted by algorithm as the degraded image.
For image enhancement problem, however, 
the original image is the degraded one.
Therefore, we employ VIF in reverse mode by specifying the enhanced
version of the image as the reference and the original image as
the degraded image.
VIF provides consistently high value of correlation between subjective MOS (Mean Opinion Score) and its scores, as shown in ~\cite{gu2016analysis}.
Besides,
it is suitable for automating evaluation process of nonlinear image enhancement algorithms~\cite{kumar2014visual}.

As shown in Table \ref{tbl:vif},
our algorithm outperforms the others in all datasets.
This means that our algorithm can maintain the visual information of images well.





\subsection{Time Cost}


Fig. \ref{fig:time} gives the comparison among different methods in terms of time cost. 
Although SRIE and NPE produce small distortion,
they are quite time-consuming.
Our method achieves the smallest distortion than others with an acceptable time cost.

\subsection{Subjective Evaluation}

Fig. \ref{fig:more} shows more examples for visual comparison.
Although the color correction post-processing in MSRCR 
can handle the color cast in some cases (\eg underwater and hazy images),
it may make the results look whitish. Besides, in some dark 
show halo artifacts around sharp edges.
The results of MSRCR shows severe halo artifacts around sharp edges and obvious noise in very dark areas (see the two astronauts) and suffer from detail loss in some bright areas (see the Christmas hat). 
The results of Dong is noisy and full of bold edges making it look like exaggerated art paintings. The results of LIME is so bright that many bright areas are saturated. Also, the noise in dark areas are amplified and de-noising method are therefore required to obtain better results.
MF may introduce color over-enhancement (see the ground beneath the astronauts' feet) and SRIE may produce slight halo effects in some edges (see the Christmas hat).




\section{Limitation And Future Work} 

\begin{figure} [t]
\centering
\subfloat[]{\includegraphics[width=1in]{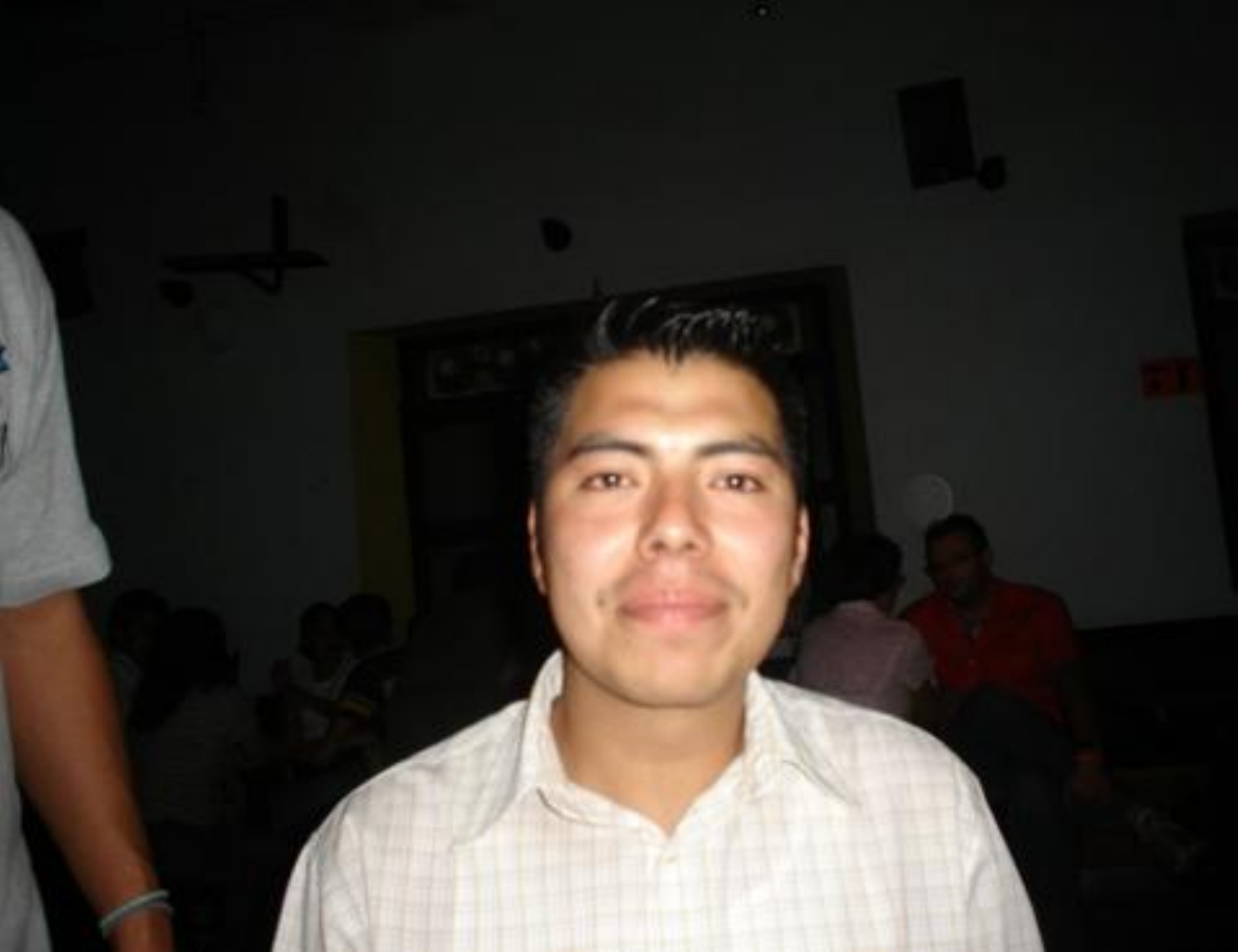} }\hfil
\subfloat[]{\includegraphics[width=1in]{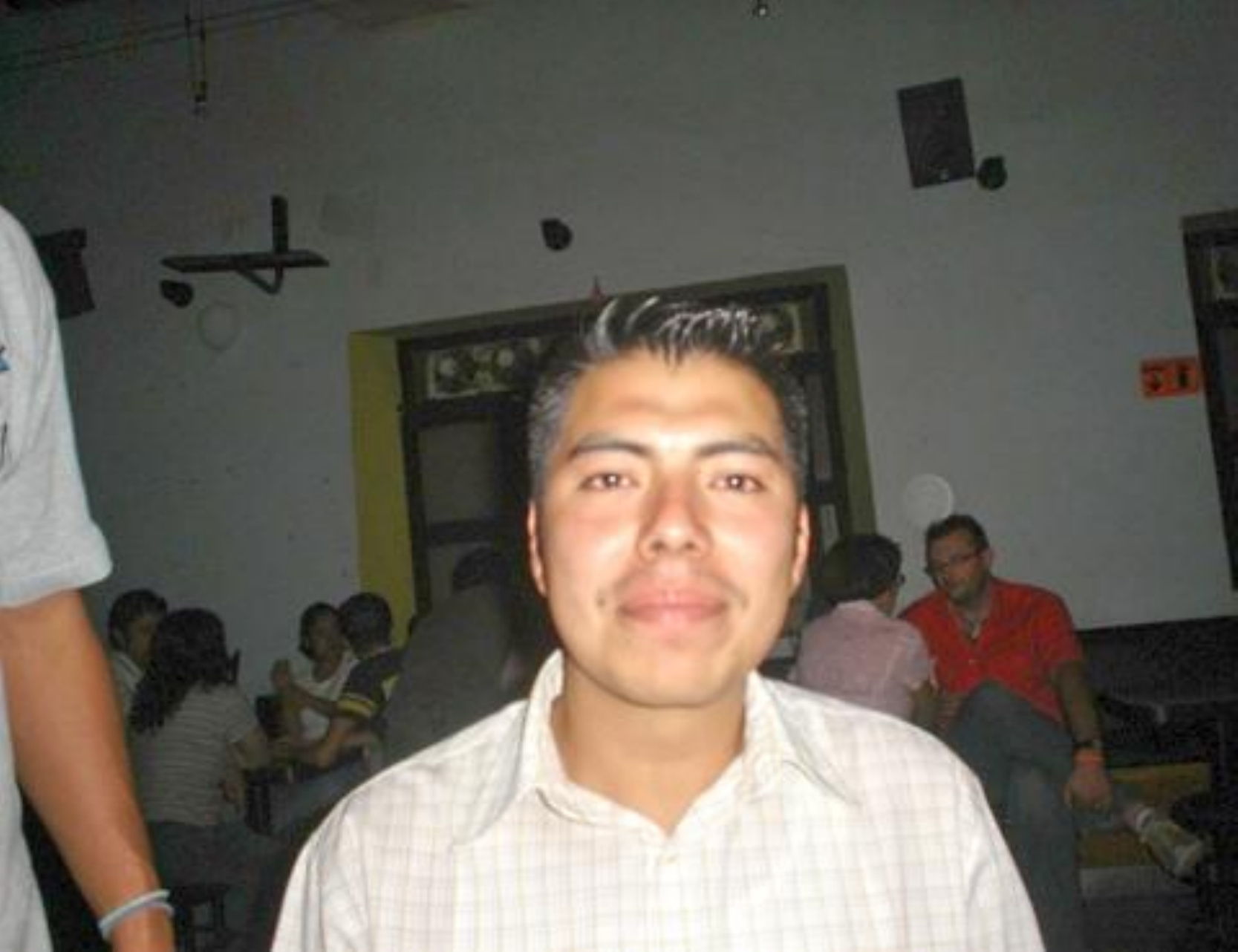} }\hfil
\subfloat[]{\includegraphics[width=1in]{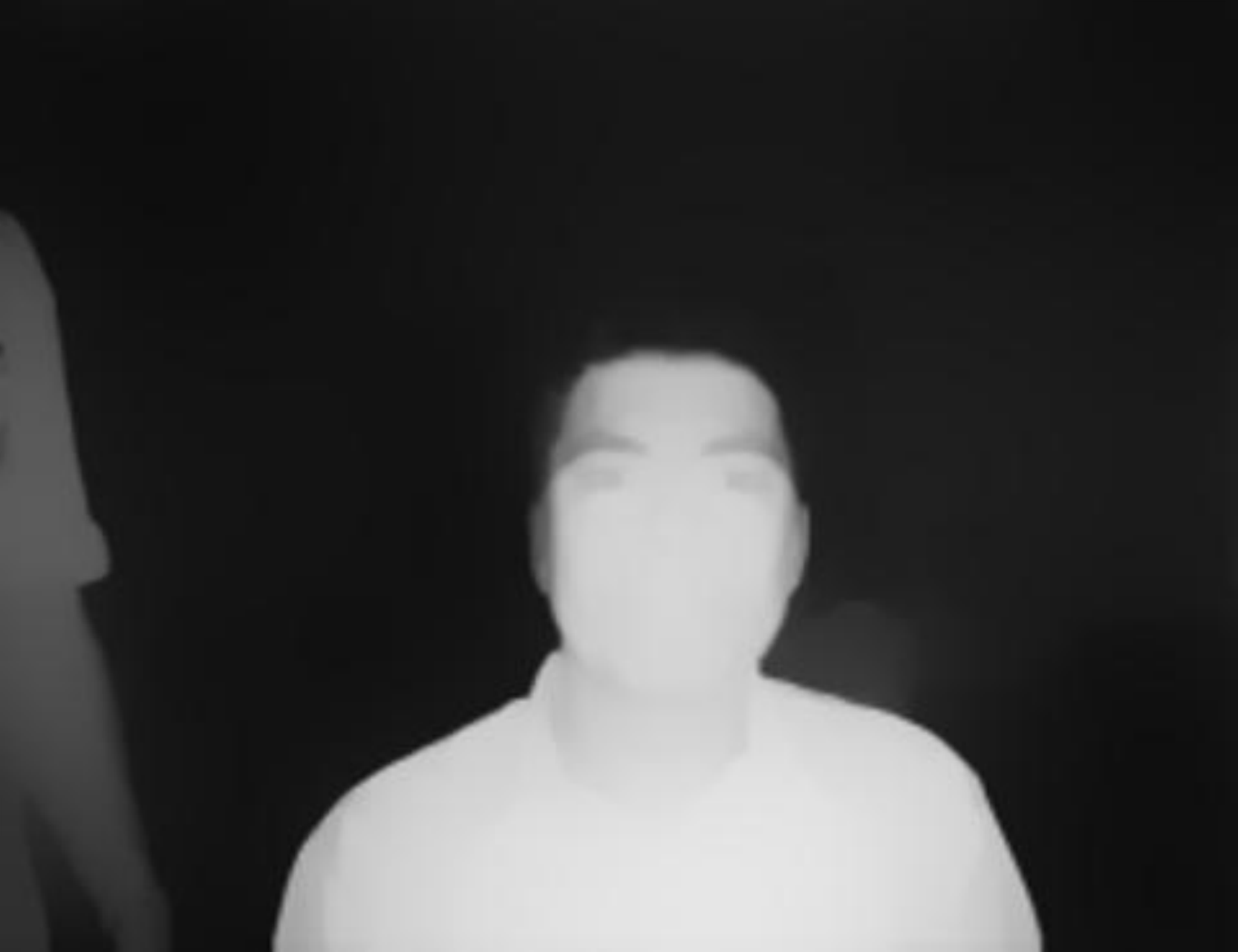} }
\caption{
An example of a failure case. 
(a) Raw image. (b) Enhanced result. (c) Estimated illumination $\textbf T$.
} 
\label{fig:failure}
\end{figure}

Fig. \ref{fig:failure} shows a failure case of our technique that
the hair of the man turns to be grey because of over-enhancement.  
This is due to the dark area behind his head blending with his black hair. 
As shown in Fig. \ref{fig:failure} (c), the hair is mistaken as the dark background in the estimated illumination map and therefore is enhanced along with the background.
Such mistake is a result of the existing illumination map estimation techniques.
This highlights a direction for future work.
To avoid the over-enhancement due to the ignorance of the scene content, 
semantic understanding is required. With further refinement, 
we might employ the deep learning techniques to estimate the illumination map.

Besides, we only use two images to obtain the enhanced result.
The over-exposure problem is remain unsolved.
Images with smaller exposures than the input image should be considered in our framework to obtain a better result. We will address this problem as future work.

\section{Conclusion}

In this paper, we propose an exposure fusion framework and an enhancement algorithm to provide an accurate contrast enhancement.
Based on our framework, we solve three problems: 1) We borrow the illumination estimation techniques to obtain the weight matrix for image fusion. 
2) We introduce our camera response model to synthesize multi-exposure images.
3) We find the best exposure ratio so that the synthetic image
is well-exposed in the regions where the original image under-exposed.
The final enhanced result is obtained by fusing the input image and the synthetic image according to the weight matrix.
The experimental results have revealed the advance of our method compared with several state-of-the-art alternatives. 
To encourage future works and allow more experimental verification and comparisons,
we make the source code open on our project website\footnote{\url{https://baidut.github.io/BIMEF}.},
as well as the relevant test code used to reproduce the experiment in the paper.
We also provide the results of other competitors to facilitate the validation of quality measures.

\appendix[Derivation of Camera Response Model] 

\input{appendix_crf_model}
\ifCLASSOPTIONcaptionsoff
  \newpage
\fi

\bibliographystyle{IEEEtran}
\end{document}

%% file: def.tex
\newcommand*{\argmax}{\operatornamewithlimits{argmax}\limits}

\newcommand{\figR}[1]{Fig.~\ref{fig:#1}}
\newcommand{\figLC}[2]{
        \caption{#2}
        \label{fig:#1}
}

\newcommand{\eqL}[1]{\label{eq:#1}}
\newcommand{\eqR}[1]{Eq.~\ref{eq:#1}}

\ifdefined\etal 	\else \newcommand{\etal}{\emph{et al. }} 	\fi
\ifdefined\eg 		\else \newcommand{\eg}{\emph{e.g. }} 		\fi
\ifdefined\ie 		\else \newcommand{\ie}{\emph{i.e. }} 		\fi
\ifdefined\etc 		\else \newcommand{\etc}{\emph{etc. }} 		\fi

\usepackage{graphicx} 
\graphicspath{ {images/} }

\usepackage{lipsum}

\usepackage{currfile-abspath}


%% file: appendix_crf_model.tex


The CRF model can be derived by solving the following comparametric equation (plug $g = \beta f^ \gamma$ to $f(kE) = g(f(E), k)$:
\begin{equation}
f(kE) = \beta f(E)^\gamma.
\end{equation}
Since $\beta$ is positive, we can take the logarithm of both sides of \eqR{our_comparametric}:
\begin{equation}
\label{eq_ln}
 \ln f(kE) = \ln \beta + \gamma \ln f(E).
\end{equation} 
%
Differentiate with respect to $E$, we get
\begin{equation}
  k \frac{ f^\prime(kE) } { f(kE) } = \gamma \frac{ f^\prime(E) } { f(E) }.
\eqL{hprime}
\end{equation}
The power function $\frac{ f^\prime(kE) } { f(kE) } = C_1 E^{a -1}$ where $a = \log_{k}{\gamma}$ can be a solution of \eqR{hprime}. Then $f$ can be solved by integration as
%
\begin{equation}
  f(E) =
  \begin{cases}
    e ^ { C_1 \frac{E^a}{a} + C_2}, & \text{if } \gamma \neq 1, \\
    C_3 E ^ {C_1 } , & \text{if } \gamma = 1. 
  \end{cases}
\end{equation}
The constants
$C_1$, $C_2$ and $C_3$ can be determined by the restriction of $f(1) = 1$ and $f(k) = \beta$ (plug $f(1) = 1$ to \eqR{our_comparametric}) 
%
%
and the resulting $f$ is obtained as follows:
\begin{equation}
  f(E) = 
  \begin{cases}
    e^{b(1 - E^a) }, & \text{if } \gamma \neq 1, \\
    E^c, & \text{if } \gamma = 1, 
  \end{cases}
\end{equation}
where $a$ and $b$ are model parameters in the case of $\gamma \neq 1$:
\begin{equation}
  a = \log_{k}{\gamma}, \quad b =\frac{\ln \beta}{1-\gamma};
\end{equation}
And $c$ is model parameter in the case of $\gamma = 1$:
\begin{equation}
  c = \log_k \beta.
\end{equation}